\documentclass[10pt, a4paper]{article}
\usepackage{lrec-coling2024}

\usepackage{epstopdf}
\usepackage[utf8]{inputenc}
\usepackage[american]{babel}
\usepackage{csquotes}

\usepackage{hyperref}
\usepackage{xstring}

\usepackage{datatool}
\usepackage{pgfplots, pgfplotstable}
\pgfplotsset{compat=newest}
\usepackage{array}
\usepackage{colortbl}
\usepackage{booktabs}
\usepackage{verbatimbox}

\usepackage{tabularx}
\usepackage{tikz}
\usepackage{longtable,supertabular}
\usepackage{booktabs} 

\title{FFSTC: Fongbe to French Speech Translation Corpus}

\newcommand{\up}[1]{\textsuperscript{#1}}
\name{D. Fortuné Kponou\up{1}, Fréjus A. A. Laleye\up{2}, Eugène C. Ezin\up{1}
}

\address{\small{1} Institut de Mathématiques et de Sciences Physiques, Dangbo, Bénin \\ 
\small{2} OPSCIDIA, Paris, France\\
\{fortune.kponou,eugene.ezin\}@imsp-uac.org,\\ frejus.laleye@opscidia.com\\}

\abstract{
In this paper, we introduce the Fongbe to French Speech Translation Corpus (FFSTC) for the first time. This corpus encompasses approximately 31 hours of collected Fongbe language content, featuring both French transcriptions and corresponding Fongbe voice recordings. FFSTC represents a comprehensive dataset compiled through various collection methods and the efforts of dedicated individuals. Furthermore, we conduct baseline experiments using Fairseq's transformer\_s and conformer models to evaluate data quality and validity. Our results indicate a score of $8.96$ for the transformer\_s model and $8.14$ for the conformer model, establishing a baseline for the FFSTC corpus.
 \\ \newline \Keywords{Speech translation corpus, spoken language translation, Low-resource, Fongbe-French, Fongbe} }

\begin{document}

\maketitleabstract

\section{Introduction}

In the era of global communication and rapid technological advancement, the development of efficient  translation systems holds immense importance. These communications systems have the potential to break language barriers, facilitating meaningful knowledge exchange, and increasing the cross-cultural connections among individuals from diverse linguistic backgrounds \cite{translationImpact}. They play a pivotal role in enhancing activities such as tourism, contributing to economic growth \cite{lenba}. Translation systems help people to access educational resources and information in their native languages, promoting literacy and knowledge sharing. \newline
While text-based translation systems have revolutionized communication for many languages, they fall short for tonal languages like Fongbe,  which are more spoken than written. This creates a critical need for speech-to-text translation tailored to tonal languages such as Fongbe. Fongbe is the most spoken dialect of Benin, by more than $50\%$ of Benin’s population, including 8 million speakers. Fongbe is also spoken in Nigeria and Togo \cite{7733280}. \newline
Speech translation involves the transformation of spoken language into text in another language. It is worth noting that speech translation has predominantly focused on languages with abundant linguistic resources, including English, French, Chinese, and Spanish \cite{ccb197e059de4d9ea06c351aa3c60bb6,jia-etal-2022-cvss,9054626}. The traditional speech translation approach called cascade method, involve two separate modules, the first module for automatic speech recognition (ASR) and the second for text translation \cite{app12031097}. This approach require two separate corpora to train individually each modules that are subsequently coupled. The advent of sequence-to-sequence architectures \cite{sutskever2014sequence} has reshaped this landscape, enabling the creation of a single corpus containing audio recordings and target texts. This development is particularly relevant for low-resource languages like Fongbe, which are in the category of spoken language and have limited written resources. So it is challenging to get written text in Fongbe and its corresponding in French, than getting a recording of voice in Fongbe with transcription in French. \newline
In response to these challenges and opportunities, we present the Fongbe-French Speech Translation Corpus (FFSTC), a unique resource comprising $31$ hours of spoken Fongbe paired with French text. Our objective is to advance voice translation technologies for less commonly studied languages like Fongbe. \newline
This paper makes two key contributions. The first is the introduction of the Fongbe-French Speech Translation Corpus (FFSTC), the first public dataset of its kind. This rich dataset, contain 31 hours of speech data with diverse speakers and topics. The second is the establishment of a baseline performance by evaluating a transformer\_s and conformer model of Fairseq toolkit \citep{ott2019fairseq}  on the FFSTC, achieving a baseline BLEU of $8.96$ and $8.14$ respectively. This benchmark provides a reference point for future research using the FFSTC and helps identify areas for improvement.\newline
This paper is structured as follows. Section \ref{sec:Existing-end-to-end-SLT-Corpora}  provides an overview of existing related studies. Section \ref{sec:Corpus-Creation-Methodology} delves into the methodology used to create the FFSTC dataset, detailing the data collection and processing procedures. Section \ref{sec:Statistics-of-the-corpus} then explores the structure and statistics of the corpus, offering insights into its composition and characteristics. Section \ref{sec:Speech-Translation-Baseline} presents the experiments conducted to establish a baseline performance for a transformer and conformer model on the FFSTC, using BLEU metrics to assess its effectiveness. Finally, Section \ref{sec:conclusion} concludes the paper by summarizing the key contributions and outlining potential future directions.

\section{Related work}
\label{sec:Existing-end-to-end-SLT-Corpora}
Large-scale speech translation datasets play a crucial role in advancing research in speech translation. However, the availability of such datasets is limited, especially for low-resource languages. To achieve meaningful breakthroughs in speech translation, it is crucial to acknowledge that, even though the research in this domain is relatively recent, substantial efforts must be focused on the construction of parallel datasets. Recent available datasets encompass a variety of language pairs, including Chinese-Mongolian\cite{ccb197e059de4d9ea06c351aa3c60bb6} , Chinese-English\cite{zhang2021bstc}, translations from 21 languages into English \cite{jia-etal-2022-cvss}, and translations from English into 15 languages \cite{wang2020covost}. The introduced corpora, such as GigaST \cite{ye2022gigast}, LibriVox \cite{beilharz-etal-2020-librivoxdeen}, LibriSpeech-FR \cite{kocabiyikoglu-etal-2018-augmenting}, mintzai-ST \cite{etchegoyhen21_iberspeech}, Multilingual TEDx Corpus \cite{Salesky2021TheMT}, and Europarl-ST \cite{9054626}, contribute to filling the gap of limited resources available for training end-to-end speech translation models. These datasets enable researchers to develop more robust and effective systems for various well-resourced languages. It is therefore undeniable that speech translation research has been extensively conducted for major languages like English, Japanese, and Spanish. However, there is a lack of research in speech translation for under-resourced languages.  Some efforts have nevertheless been made for the creation of resources for certain low-resource languages. 
\citet{10.1007/978-3-319-95153-9_12}, have created an Amharic speech corpus by preparing $7.43$ hours of read-speech from $8,112$ sentences. Additionally, they have developed a parallel Amharic-English corpus of $19,972$ sentences with tourism as the application domain. Another study focuses on constructing a large corpus for speech translation from Khmer (Cambodian) to English and French \cite{9660421}. The corpus includes approximately $155$ hours of speech and $1.7$ million words of text from the Extraordinary Chambers in the Courts of Cambodia (ECCC). 
Moreover, another project aims to provide datasets for Tamasheq, a developing language spoken in Mali and Niger \cite{boito2022speech}. The datasets consist of radio recordings from Studio Kalangou in Niger and Studio Tamani in Mali. They include a large amount of unlabeled audio data in five languages (French, Fulfulde, Hausa, Tamasheq, and Zarma) and a smaller $17$-hour parallel corpus of audio recordings with translations in French. The quantity of data collected for the Ahmaric and Tamasheq languages, compared to other well-resourced languages, confirms the difficulty in establishing solid dataset for a system of speech translation for low-resourced languages. For these languages, the lack of available speech and text corpora remains a significant challenge for speech translation.

\section{Methodology}
\label{sec:Corpus-Creation-Methodology}
In this section, we delve into the intricate process of creating the corpus, shedding light on the meticulous steps taken to acquire and compile the translations. Our methodology not only outlines the data collection procedure but also discusses the quality control measures implemented to ensure the reliability of the dataset.

\subsection{Clips creation process}
The data within this corpus originates from three distinct sources, each contributing significantly to its composition. The first source is the ALFFA corpus \citep{laleye2016first}, from which we extracted sentences in the Fongbe language and corresponding audio recording. These sentences were subsequently translated into French by experienced linguists.

The second source is the FFR1.1 corpus \citep{dossou2020ffr}, which serves as a corpus for machine translation. It comprises sentences in French, accompanied by their written translations in the Fongbe language. In this case, we provided the Fongbe sentences to linguists who read and recorded them through a dedicated web platform.

Finally, the third source involves an assembly of sentences gathered from various books, including "Kondo le requin" \citep{pliya-2013} and "Un piège sans fin" \citep{bellhy}. We also excerpts from texts found on news websites. These sentences were thoughtfully selected, condensed, and made accessible via a web platform. Given the complexity of this task, we encouraged participants to collaborate within teams to ensure the coherence and quality of the collection process.

The participants teams provided Fongbe translations for the given sentences in French. They engaged in a voice recording process using laptops or smartphones through our web platform's interface. This involved audibly translating the sentences displayed in French on the screen. Following this, recorded submissions underwent a rigorous validation process by designated validators. These validators used a straightforward voting system to assess and determine the quality of each translation.

\subsection{Clips validation process}
In the case of the first source, French transcripts derived from Fongbe transcriptions undergo a meticulous peer-review process by a second validator. This stringent review ensures the translation's accuracy and conformity.

For the second source, the validation process adopts a straightforward binary approach, utilizing a yes-or-no voting system. Audio recordings that align precisely with the provided Fongbe transcripts are retained, while those that do not are omitted.

The third source follows a selective inclusion process. Only clips endorsed as valid by the system are integrated into the corpus. Evaluation criteria encompass multiple facets, including translation completeness, audio recording clarity, and overall translation quality. Our platform features a robust voting system that facilitates the filtration of clips created by participants based on validators' assessments. The voting system employs a scale ranging from zero to five. Recordings that receive a score below two are classified as invalid and are consequently excluded from the dataset. Conversely, recordings scoring three or higher are regarded as valid contributions and are included in the dataset for further analysis.
Subsequently, all recordings undergo a comprehensive review by a dedicated team of validators, who make the final determination of approval or rejection.

\section{Corpus structure and statistics }
\label{sec:Statistics-of-the-corpus}

In the previous section of the report, we outlined the origins of our dataset, which emanate from three distinct sources. The ASR corpus ALFFA \citep{laleye2016first} translation, a recording selection of sentences extracted from the FFR1.1 corpus \citep{dossou2020ffr} with recorded readings, and a collaborative effort within a team for translating French sentences.

Combining the first two sources, we accumulated a total of 10 hours of data, while the latter made a significant contribution of 21 hours. The team work was done by a group of volunteers who were very committed. There were 8 women and 12 men in the group, all between 20 and 40 years old, making a total of 20 people. 

Following the corpus collection, we splited it into training (Train), development (Dev), and test (Test) sets, as outlined in Table \ref{table:stats}. The corpus, in its entirety, comprises 16,447 sentences, which were presented randomly to the participants. Additionally, multiple participants within each team were allowed to record simultaneously using the same credentials.

\begin{figure}[!htbp]
    \begin{tikzpicture}
        \begin{axis}[
            ybar,
            ymin=0,
            ylabel={Count},
            xlabel={Split},
            symbolic x coords={Train, Dev, Test},
            xtick=data,
            legend style={at={(0.5,-0.15)}, anchor=north, legend columns=-1},
        ]
        \addplot table[x=Split,y=Vocabulary] {
            Split Vocabulary
            Train 7476
            Dev 4277
            Test 3647
        };
        \addplot table[x=Split,y=Words] {
            Split Words
            Train 104213
            Dev 39564
            Test 25751
        };
        \end{axis}
    \end{tikzpicture}
    \caption{Vocabulary and number of words of each split}
    \label{fig:vocabulary}
\end{figure}
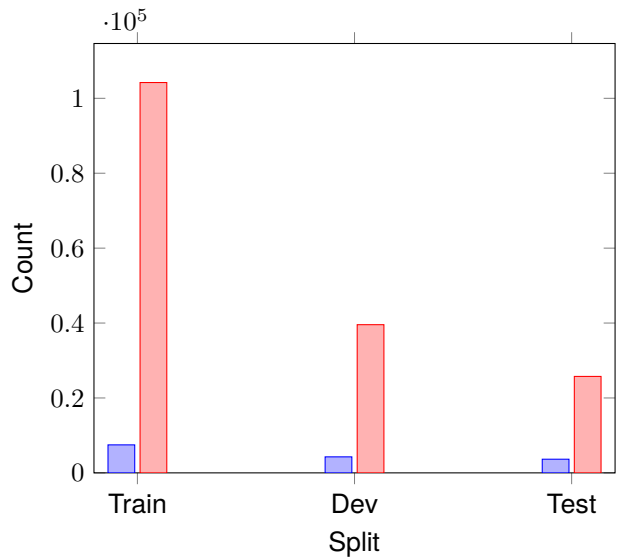

Figure \ref{fig:vocabulary} illustrates the distribution of vocabulary and word counts across the Train, Dev, and Test partitions of our collected dataset for the Fongbe speech translation task. Figure \ref{fig:vocabulary} highlights the substantial vocabulary diversity and word richness in our dataset, with Train exhibiting the highest vocabulary count at approximately 7,500 unique words. This variation in vocabulary and word counts across splits underscores the dataset's suitability for training and evaluating speech translation models with varying language complexities.

\begin{table}[h]
\centering
\begin{tabular}{lcc}
\hline
Split & Sentences & Audio (hours) \\
\hline
Train & 11636 & 20 \\
Dev & 2329 & 7 \\
Test & 2482 & 4 \\
\hline
\multicolumn{3}{l}{\textbf{Minimum sentence length (words)}} \\
Train & 1 & - \\
Dev & 1 & - \\
Test & 1 & - \\
\multicolumn{3}{l}{\textbf{Maximum sentence length (words)}} \\
Train & 136 & - \\
Dev & 45 & - \\
Test & 63 & - \\
\hline
\end{tabular}
\caption{Corpus statistics.}
\label{table:stats}
\end{table}

\section{Fongbe-French End-to-End Speech Translation Baseline}
\label{sec:Speech-Translation-Baseline}
In this work, we use the Fairseq toolkit to establish a baseline for Fongbe-French End-to-End speech translation. We run the baseline experiments using a state-of-the-art approach, implementing the transformer\_s and conformer architecture within Fairseq framework to build a speech translation model. 

\subsection{Experiments}
We ran the experiments on the collected dataset with the downsampling of the initially recorded audio clips from 44.1 kHz to 16 kHz during the preprocessing step. Following this, to maintain uniformity and streamline the training process, we subjected the extracted features to a normalization procedure, aligning them to a mean of 0 and a standard deviation of 1. As part of our data quality control measures, we implemented a filtering step to exclude samples that exceeded $3,000$ frames, thus ensuring the quality and manageability of the dataset. 

We used for 80-dimensional mel-filter bank features and built character-level vocabulary using Sentencepiece \citep{kudo2018sentencepiece}. We employed a label-smoothed cross-entropy loss function with a label-smoothing factor set to $0.1$. Optimizing the models utilized the Adam optimizer with a learning rate of $1e-3$, coupled with a learning rate scheduler of the inverse square root type, featuring warm-up updates and gradient clipping. The model is trained for 500 epochs using the Adam optimizer \citep{kingma2014adam}. 





\subsection{Results and discussion}

We utilized a beam size of $5$ and averaged the last $10$ checkpoints when evaluating the baseline model trained on the test set. Our evaluation metric was case-insensitive tokenized BLEU, implemented using the sacreBLEU toolkit \citep{post2018call} following the methodology proposed by \citep{papineni-etal-2002-bleu}.

\begin{table}[!ht]
    \centering
    \begin{tabularx}{\columnwidth}{l l X}
        \hline
         Architecture & enc-type & FFSTC (test) \\
         \hline
         transformer\_s & - &8.96 (18.4/10.9/10.7/10.7) \\ 
         \hline
         conformer & abs & 8.14 (16.3/8.8/8.5/8.4) \\ 
         \hline
    \end{tabularx}
    \caption{Baseline for Fongbe-French End-to-end speech translation BLEU-4 results.}
    \label{tab:scores}
\end{table}

The BLEU score achieved on the test set of the collected dataset using the transformer\_s and conformer architecture, is reported in Table \ref{tab:scores}. The reported score is dissected into precision scores for various n-gram lengths. For the transformer\_s there is $28.7\%$ for unigrams (1-grams), $4.6\%$ for bigrams (2-grams), $1.8\%$ for trigrams (3-grams), and $0.6\%$ for 4-grams. For the conformer there is  $16.3$ for 1-grams, $8.8$ for 2-grams, $8.5$ for 3-grams and $8.4$ for 4-grams. We set the positional encoding for the conformer to absolute.
This evaluation provides valuable insights into the quality of our collected dataset. Despite the relatively low BLEU score of $8.96$ for the transformer\_s and $8.14$ for the conformer, which signifies room for improvement, it is worth noting that this score holds promise for the development of an effective Fongbe-French end-to-end speech translation model. While there is certainly room for enhancement, particularly in terms of n-gram precision, this result serves as a foundation upon which further refinements and optimizations can be built.

In future research endeavors, we anticipate that the integration of pre-training models and advanced techniques, such as transfer learning, could significantly enhance the performance of the baseline model. Leveraging the rich resources of pre-training models in French and further optimization strategies promises to elevate the accuracy and fluency of Fongbe-French speech translation, thus bridging linguistic and cultural divides more effectively. This work lays the groundwork for future studies aimed at harnessing the full potential of state-of-the-art models and data-driven approaches to advance the field of speech translation for underrepresented languages like Fongbe.

\section{Conclusion}
\label{sec:conclusion}
In this paper, we introduced FFTSC, the first speech translation corpus for Fongbe language. This dataset has been carefully curated, including a wide variety of language details and difficulties for tonal language. It is incredibly valuable for researchers and professionals who are working on speech translation and natural language processing. It stands as the first of its kind for the West African languages, contributing not only to the advancement of Fongbe-French translation but also to the broader goal of preserving and promoting linguistic diversity. Our baseline experiments, conducted on this novel resource, demonstrate promising potential for the development of robust speech translation systems. We envision that this dataset will play a pivotal role in bridging linguistic and cultural divides, facilitating enhanced communication and mutual understanding across languages and communities in the West African context. Its availability paves the way for exciting future research endeavors and applications in the field.

\section{Acknowledgements}
We would like to thank the Partnership for Skills in Applied Sciences, Engineering, and Technology (PASET) through the Regional Scholarship and Innovation Fund (RSIF) for the support for this research.

\nocite{*}

\bibliographystyle{lrec-coling2024-natbib}
\bibliography{lrec-coling2024-example}

\begin{thebibliography}{0}
\expandafter\ifx\csname natexlab\endcsname\relax\def\natexlab#1{#1}\fi

\end{thebibliography}


\begin{thebibliography}{46}
\expandafter\ifx\csname natexlab\endcsname\relax\def\natexlab#1{#1}\fi

\bibitem[{Aho and Ullman(1972)}]{Aho:72}
Alfred~V. Aho and Jeffrey~D. Ullman. 1972.
\newblock \emph{The Theory of Parsing, Translation and Compiling}, volume~1.
\newblock Prentice-Hall, Englewood Cliffs, NJ.

\bibitem[{Anwar et~al.(2023)Anwar, Shi, Goswami, Hsu, Pino, and
  Wang}]{anwar2023muavic}
Mohamed Anwar, Bowen Shi, Vedanuj Goswami, Wei-Ning Hsu, Juan Pino, and
  Changhan Wang. 2023.
\newblock Muavic: A multilingual audio-visual corpus for robust speech
  recognition and robust speech-to-text translation.
\newblock \emph{arXiv preprint arXiv:2303.00628}.

\bibitem[{Bang et~al.(2023{\natexlab{a}})Bang, Maeng, Park, Yun, and
  Kim}]{bang2023english}
Jeong-Uk Bang, Joon-Gyu Maeng, Jun Park, Seung Yun, and Sang-Hun Kim.
  2023{\natexlab{a}}.
\newblock English--korean speech translation corpus (enkost-c): Construction
  procedure and evaluation results.
\newblock \emph{ETRI Journal}, 45(1):18--27.

\bibitem[{Bang et~al.(2023{\natexlab{b}})Bang, Maeng, Park, Yun, and
  Kim}]{WOS:000811600300001}
Jeong-Uk Bang, Joon-Gyu Maeng, Jun Park, Seung Yun, and Sang-Hun Kim.
  2023{\natexlab{b}}.
\newblock \href {https://doi.org/10.4218/etrij.2021-0336} {English-korean
  speech translation corpus (enkost-c): Construction procedure and evaluation
  results}.
\newblock \emph{ETRI JOURNAL}, 45(1):18--27.

\bibitem[{Beilharz et~al.(2020)Beilharz, Sun, Karimova, and
  Riezler}]{beilharz-etal-2020-librivoxdeen}
Benjamin Beilharz, Xin Sun, Sariya Karimova, and Stefan Riezler. 2020.
\newblock \href {https://aclanthology.org/2020.lrec-1.441}
  {{L}ibri{V}ox{D}e{E}n: A corpus for {G}erman-to-{E}nglish speech translation
  and {G}erman speech recognition}.
\newblock In \emph{Proceedings of the Twelfth Language Resources and Evaluation
  Conference}, pages 3590--3594, Marseille, France. European Language Resources
  Association.

\bibitem[{Boito et~al.(2022)Boito, Bougares, Barbier, Gahbiche, Barrault,
  Rouvier, and Est\'eve}]{boito2022speech}
Marcely~Zanon Boito, Fethi Bougares, Florentin Barbier, Souhir Gahbiche, Lo\"ic
  Barrault, Mickael Rouvier, and Yannick Est\'eve. 2022.
\newblock Speech resources in the tamasheq language.
\newblock \emph{Language Resources and Evaluation Conference (LREC)}.

\bibitem[{Cho et~al.(2021)Cho, Kim, Cho, and Kim}]{cho2021kosp2e}
Won~Ik Cho, Seok~Min Kim, Hyunchang Cho, and Nam~Soo Kim. 2021.
\newblock Kosp2e: Korean speech to english translation corpus.
\newblock \emph{arXiv preprint arXiv:2107.02875}.

\bibitem[{Dadi{\'e}(1955)}]{dadié1955pagne}
B.B. Dadi{\'e}. 1955.
\newblock \href {https://books.google.bj/books?id=e4EdAQAAIAAJ} {\emph{Le pagne
  noir: contes africains}}.
\newblock Pr{\'e}sence africaine: Ed. Africaines. Pr{\'e}sence africaine.

\bibitem[{Di~Gangi et~al.(2019)Di~Gangi, Cattoni, Bentivogli, Negri, and
  Turchi}]{di2019must}
Mattia~A Di~Gangi, Roldano Cattoni, Luisa Bentivogli, Matteo Negri, and Marco
  Turchi. 2019.
\newblock Must-c: a multilingual speech translation corpus.
\newblock In \emph{Proceedings of the 2019 Conference of the North American
  Chapter of the Association for Computational Linguistics: Human Language
  Technologies, Volume 1 (Long and Short Papers)}, pages 2012--2017.
  Association for Computational Linguistics.

\bibitem[{Doherty(2016)}]{translationImpact}
Stephen Doherty. 2016.
\newblock The impact of translation technologies on the process and product of
  translation.
\newblock \emph{International Journal of Communication}, 10:969.

\bibitem[{Dossou and Emezue(2020)}]{dossou2020ffr}
Bonaventure~FP Dossou and Chris~C Emezue. 2020.
\newblock Ffr v1. 1: Fon-french neural machine translation.
\newblock \emph{arXiv preprint arXiv:2006.09217}.

\bibitem[{Etchegoyhen et~al.(2022)Etchegoyhen, Arzelus, Gete, Alvarez, Torre,
  Martín-Doñas, González-Docasal, and Fernandez}]{app12031097}
Thierry Etchegoyhen, Haritz Arzelus, Harritxu Gete, Aitor Alvarez, Iván~G.
  Torre, Juan~Manuel Martín-Doñas, Ander González-Docasal, and Edson~Benites
  Fernandez. 2022.
\newblock \href {https://doi.org/10.3390/app12031097} {Cascade or direct speech
  translation? a case study}.
\newblock \emph{Applied Sciences}, 12(3).

\bibitem[{Etchegoyhen et~al.(2021)Etchegoyhen, Arzelus, {Gete Ugarte}, Alvarez,
  González-Docasal, and {Benites Fernandez}}]{etchegoyhen21_iberspeech}
Thierry Etchegoyhen, Haritz Arzelus, Harritxu {Gete Ugarte}, Aitor Alvarez,
  Ander González-Docasal, and Edson {Benites Fernandez}. 2021.
\newblock \href {https://doi.org/10.21437/IberSPEECH.2021-41} {{mintzai-ST:
  Corpus and Baselines for Basque-Spanish Speech Translation}}.
\newblock In \emph{Proc. IberSPEECH 2021}, pages 190--194.

\bibitem[{Federmann and Lewis(2016)}]{federmann-lewis-2016-microsoft}
Christian Federmann and William~D. Lewis. 2016.
\newblock \href {https://aclanthology.org/2016.iwslt-1.12} {{M}icrosoft speech
  language translation ({MSLT}) corpus: The {IWSLT} 2016 release for {E}nglish,
  {F}rench and {G}erman}.
\newblock In \emph{Proceedings of the 13th International Conference on Spoken
  Language Translation}, Seattle, Washington D.C. International Workshop on
  Spoken Language Translation.

\bibitem[{Gulati et~al.(2020)Gulati, Qin, Chiu, Parmar, Zhang, Yu, Han, Wang,
  Zhang, Wu et~al.}]{gulati2020conformer}
Anmol Gulati, James Qin, Chung-Cheng Chiu, Niki Parmar, Yu~Zhang, Jiahui Yu,
  Wei Han, Shibo Wang, Zhengdong Zhang, Yonghui Wu, et~al. 2020.
\newblock Conformer: Convolution-augmented transformer for speech recognition.
\newblock \emph{arXiv preprint arXiv:2005.08100}.

\bibitem[{Iranzo-S{\'a}nchez et~al.(2020)Iranzo-S{\'a}nchez, Silvestre-Cerda,
  Jorge, Rosell{\'o}, Gim{\'e}nez, Sanchis, Civera, and
  Juan}]{iranzo2020europarl}
Javier Iranzo-S{\'a}nchez, Joan~Albert Silvestre-Cerda, Javier Jorge, Nahuel
  Rosell{\'o}, Adria Gim{\'e}nez, Albert Sanchis, Jorge Civera, and Alfons
  Juan. 2020.
\newblock Europarl-st: A multilingual corpus for speech translation of
  parliamentary debates.
\newblock In \emph{ICASSP 2020-2020 IEEE International Conference on Acoustics,
  Speech and Signal Processing (ICASSP)}, pages 8229--8233. IEEE.

\bibitem[{Iranzo-Sánchez et~al.(2020)Iranzo-Sánchez, Silvestre-Cerdà, Jorge,
  Roselló, Giménez, Sanchis, Civera, and Juan}]{9054626}
Javier Iranzo-Sánchez, Joan~Albert Silvestre-Cerdà, Javier Jorge, Nahuel
  Roselló, Adrià Giménez, Albert Sanchis, Jorge Civera, and Alfons Juan.
  2020.
\newblock \href {https://doi.org/10.1109/ICASSP40776.2020.9054626}
  {Europarl-st: A multilingual corpus for speech translation of parliamentary
  debates}.
\newblock In \emph{ICASSP 2020 - 2020 IEEE International Conference on
  Acoustics, Speech and Signal Processing (ICASSP)}, pages 8229--8233.

\bibitem[{Jia et~al.(2022)Jia, Tadmor~Ramanovich, Wang, and
  Zen}]{jia-etal-2022-cvss}
Ye~Jia, Michelle Tadmor~Ramanovich, Quan Wang, and Heiga Zen. 2022.
\newblock \href {https://aclanthology.org/2022.lrec-1.720} {{CVSS} corpus and
  massively multilingual speech-to-speech translation}.
\newblock In \emph{Proceedings of the Thirteenth Language Resources and
  Evaluation Conference}, pages 6691--6703, Marseille, France. European
  Language Resources Association.

\bibitem[{Kingma and Ba(2014)}]{kingma2014adam}
Diederik~P Kingma and Jimmy Ba. 2014.
\newblock Adam: A method for stochastic optimization.
\newblock \emph{arXiv preprint arXiv:1412.6980}.

\bibitem[{Kocabiyikoglu et~al.(2018)Kocabiyikoglu, Besacier, and
  Kraif}]{kocabiyikoglu-etal-2018-augmenting}
Ali~Can Kocabiyikoglu, Laurent Besacier, and Olivier Kraif. 2018.
\newblock \href {https://aclanthology.org/L18-1001} {Augmenting librispeech
  with {F}rench translations: A multimodal corpus for direct speech translation
  evaluation}.
\newblock In \emph{Proceedings of the Eleventh International Conference on
  Language Resources and Evaluation ({LREC} 2018)}, Miyazaki, Japan. European
  Language Resources Association (ELRA).

\bibitem[{Kudo and Richardson(2018)}]{kudo2018sentencepiece}
Taku Kudo and John Richardson. 2018.
\newblock Sentencepiece: A simple and language independent subword tokenizer
  and detokenizer for neural text processing.
\newblock \emph{arXiv preprint arXiv:1808.06226}.

\bibitem[{Laleye et~al.(2016{\natexlab{a}})Laleye, Besacier, Ezin, and
  Motamed}]{laleye2016first}
Fr{\'e}jus~AA Laleye, Laurent Besacier, Eug{\`e}ne~C Ezin, and Cina Motamed.
  2016{\natexlab{a}}.
\newblock First automatic fongbe continuous speech recognition system:
  Development of acoustic models and language models.
\newblock In \emph{2016 Federated Conference on Computer Science and
  Information Systems (FedCSIS)}, pages 477--482. IEEE.

\bibitem[{Laleye et~al.(2016{\natexlab{b}})Laleye, Besacier, Ezin, and
  Motamed}]{7733280}
Fréjus A.~A. Laleye, Laurent Besacier, Eugène~C. Ezin, and Cina Motamed.
  2016{\natexlab{b}}.
\newblock First automatic fongbe continuous speech recognition system:
  Development of acoustic models and language models.
\newblock In \emph{2016 Federated Conference on Computer Science and
  Information Systems (FedCSIS)}, pages 477--482.

\bibitem[{Lenba and Ennebati(2022)}]{lenba}
Noureddine Lenba and Fatima~Zohra Ennebati. 2022.
\newblock The role of translation in the development of tourism in algeria :
  Obstacles and opportunities.

\bibitem[{Nguyen et~al.(2022)Nguyen, Tran, Doan, Luong, and
  Nguyen}]{nguyen2022highquality}
Linh~The Nguyen, Nguyen~Luong Tran, Long Doan, Manh Luong, and Dat~Quoc Nguyen.
  2022.
\newblock \href {http://arxiv.org/abs/2208.04243} {A high-quality and
  large-scale dataset for english-vietnamese speech translation}.

\bibitem[{Olympe(1960)}]{bellhy}
BHÊLY-QUENUM Olympe. 1960.
\newblock \emph{{Un piège sans fin}}, editions présence africaine edition.

\bibitem[{Ott et~al.(2019)Ott, Edunov, Baevski, Fan, Gross, Ng, Grangier, and
  Auli}]{ott2019fairseq}
Myle Ott, Sergey Edunov, Alexei Baevski, Angela Fan, Sam Gross, Nathan Ng,
  David Grangier, and Michael Auli. 2019.
\newblock fairseq: A fast, extensible toolkit for sequence modeling.
\newblock In \emph{Proceedings of NAACL-HLT 2019: Demonstrations}.

\bibitem[{Papineni et~al.(2002{\natexlab{a}})Papineni, Roukos, Ward, and
  Zhu}]{papineni-etal-2002-bleu}
Kishore Papineni, Salim Roukos, Todd Ward, and Wei-Jing Zhu.
  2002{\natexlab{a}}.
\newblock \href {https://doi.org/10.3115/1073083.1073135} {{B}leu: a method for
  automatic evaluation of machine translation}.
\newblock In \emph{Proceedings of the 40th Annual Meeting of the Association
  for Computational Linguistics}, pages 311--318, Philadelphia, Pennsylvania,
  USA. Association for Computational Linguistics.

\bibitem[{Papineni et~al.(2002{\natexlab{b}})Papineni, Roukos, Ward, and
  Zhu}]{papineni2002bleu}
Kishore Papineni, Salim Roukos, Todd Ward, and Wei-Jing Zhu.
  2002{\natexlab{b}}.
\newblock Bleu: a method for automatic evaluation of machine translation.
\newblock In \emph{Proceedings of the 40th annual meeting of the Association
  for Computational Linguistics}, pages 311--318.

\bibitem[{Pliya(1981)}]{pliya-2013}
Jean Pliya. 1981.
\newblock \emph{{Kondo, le requin}}, editions clé en coédition avec nena
  edition.

\bibitem[{Post(2018)}]{post2018call}
Matt Post. 2018.
\newblock A call for clarity in reporting bleu scores.
\newblock \emph{arXiv preprint arXiv:1804.08771}.

\bibitem[{Post et~al.(2013)Post, Kumar, Lopez, Karakos, Callison-Burch, and
  Khudanpur}]{post2013improved}
Matt Post, Gaurav Kumar, Adam Lopez, Damianos Karakos, Chris Callison-Burch,
  and Sanjeev Khudanpur. 2013.
\newblock Improved speech-to-text translation with the fisher and callhome
  spanish-english speech translation corpus.
\newblock In \emph{Proceedings of the 10th International Workshop on Spoken
  Language Translation: Papers}.

\bibitem[{Povey et~al.(2011)Povey, Ghoshal, Boulianne, Burget, Glembek, Goel,
  Hannemann, Motlicek, Qian, Schwarz et~al.}]{povey2011kaldi}
Daniel Povey, Arnab Ghoshal, Gilles Boulianne, Lukas Burget, Ondrej Glembek,
  Nagendra Goel, Mirko Hannemann, Petr Motlicek, Yanmin Qian, Petr Schwarz,
  et~al. 2011.
\newblock The kaldi speech recognition toolkit.
\newblock In \emph{IEEE 2011 workshop on automatic speech recognition and
  understanding}, CONF. IEEE Signal Processing Society.

\bibitem[{Qi et~al.(2022)Qi, B.Teniger, Sun, and
  Zhao}]{ccb197e059de4d9ea06c351aa3c60bb6}
Xiao~Ke Qi, Borjigin B.Teniger, Yuan Sun, and Xiaobing Zhao. 2022.
\newblock \href {https://doi.org/10.11922/sciencedb.j00001.00345} {{A dataset
  of Mongolian-Chinese speech translation}}.

\bibitem[{Salesky et~al.(2021{\natexlab{a}})Salesky, Wiesner, Bremerman,
  Cattoni, Negri, Turchi, Oard, and Post}]{inproceedingstedxcorpus}
Elizabeth Salesky, Matthew Wiesner, Jacob Bremerman, Roldano Cattoni, Matteo
  Negri, Marco Turchi, Douglas Oard, and Matt Post. 2021{\natexlab{a}}.
\newblock \href {https://doi.org/10.21437/Interspeech.2021-11} {The
  multilingual tedx corpus for speech recognition and translation}.
\newblock pages 3655--3659.

\bibitem[{Salesky et~al.(2021{\natexlab{b}})Salesky, Wiesner, Bremerman,
  Cattoni, Negri, Turchi, Oard, and Post}]{Salesky2021TheMT}
Elizabeth Salesky, Matthew Wiesner, Jacob Bremerman, Roldano Cattoni, Matteo
  Negri, Marco Turchi, Douglas~W. Oard, and Matt Post. 2021{\natexlab{b}}.
\newblock \href {https://api.semanticscholar.org/CorpusID:231786401} {The
  multilingual tedx corpus for speech recognition and translation}.
\newblock \emph{ArXiv}, abs/2102.01757.

\bibitem[{Soky et~al.(2021)Soky, Mimura, Kawahara, Li, Ding, Chu, and
  Sam}]{9660421}
Kak Soky, Masato Mimura, Tatsuya Kawahara, Sheng Li, Chenchen Ding, Chenhui
  Chu, and Sethserey Sam. 2021.
\newblock \href {https://doi.org/10.1109/O-COCOSDA202152914.2021.9660421}
  {Khmer speech translation corpus of the extraordinary chambers in the courts
  of cambodia (eccc)}.
\newblock In \emph{2021 24th Conference of the Oriental COCOSDA International
  Committee for the Co-ordination and Standardisation of Speech Databases and
  Assessment Techniques (O-COCOSDA)}, pages 122--127.

\bibitem[{Sutskever et~al.(2014)Sutskever, Vinyals, and
  Le}]{sutskever2014sequence}
Ilya Sutskever, Oriol Vinyals, and Quoc~V Le. 2014.
\newblock Sequence to sequence learning with neural networks.
\newblock \emph{Advances in neural information processing systems}, 27.

\bibitem[{Vaswani et~al.(2017)Vaswani, Shazeer, Parmar, Uszkoreit, Jones,
  Gomez, Kaiser, and Polosukhin}]{vaswani2017attention}
Ashish Vaswani, Noam Shazeer, Niki Parmar, Jakob Uszkoreit, Llion Jones,
  Aidan~N Gomez, {\L}ukasz Kaiser, and Illia Polosukhin. 2017.
\newblock Attention is all you need.
\newblock \emph{Advances in neural information processing systems}, 30.

\bibitem[{Wang et~al.(2020{\natexlab{a}})Wang, Pino, Wu, and
  Gu}]{wang20201covost}
Changhan Wang, Juan Pino, Anne Wu, and Jiatao Gu. 2020{\natexlab{a}}.
\newblock Covost: A diverse multilingual speech-to-text translation corpus.
\newblock \emph{arXiv preprint arXiv:2002.01320}.

\bibitem[{Wang et~al.(2021)Wang, Riviere, Lee, Wu, Talnikar, Haziza,
  Williamson, Pino, and Dupoux}]{wang2021voxpopuli}
Changhan Wang, Morgane Riviere, Ann Lee, Anne Wu, Chaitanya Talnikar, Daniel
  Haziza, Mary Williamson, Juan Pino, and Emmanuel Dupoux. 2021.
\newblock Voxpopuli: A large-scale multilingual speech corpus for
  representation learning, semi-supervised learning and interpretation.
\newblock \emph{arXiv preprint arXiv:2101.00390}.

\bibitem[{Wang et~al.(2020{\natexlab{b}})Wang, Tang, Ma, Wu, Popuri, Okhonko,
  and Pino}]{wang2020fairseq}
Changhan Wang, Yun Tang, Xutai Ma, Anne Wu, Sravya Popuri, Dmytro Okhonko, and
  Juan Pino. 2020{\natexlab{b}}.
\newblock Fairseq s2t: Fast speech-to-text modeling with fairseq.
\newblock \emph{arXiv preprint arXiv:2010.05171}.

\bibitem[{Wang et~al.(2020{\natexlab{c}})Wang, Wu, and Pino}]{wang2020covost}
Changhan Wang, Anne Wu, and Juan Pino. 2020{\natexlab{c}}.
\newblock Covost 2 and massively multilingual speech-to-text translation.
\newblock \emph{arXiv preprint arXiv:2007.10310}.

\bibitem[{Woldeyohannis et~al.(2018)Woldeyohannis, Besacier, and
  Meshesha}]{10.1007/978-3-319-95153-9_12}
Michael~Melese Woldeyohannis, Laurent Besacier, and Million Meshesha. 2018.
\newblock A corpus for amharic-english speech translation: The case of tourism
  domain.
\newblock In \emph{Information and Communication Technology for Development for
  Africa}, pages 129--139, Cham. Springer International Publishing.

\bibitem[{Ye et~al.(2022)Ye, Zhao, Ko, Meng, Wang, Wang, and
  Cao}]{ye2022gigast}
Rong Ye, Chengqi Zhao, Tom Ko, Chutong Meng, Tao Wang, Mingxuan Wang, and Jun
  Cao. 2022.
\newblock Gigast: A 10,000-hour pseudo speech translation corpus.
\newblock \emph{arXiv preprint arXiv:2204.03939}.

\bibitem[{Zhang et~al.(2021)Zhang, Wang, Zhang, He, Wu, Li, Wang, Chen, and
  Li}]{zhang2021bstc}
Ruiqing Zhang, Xiyang Wang, Chuanqiang Zhang, Zhongjun He, Hua Wu, Zhi Li,
  Haifeng Wang, Ying Chen, and Qinfei Li. 2021.
\newblock \href {http://arxiv.org/abs/2104.03575} {Bstc: A large-scale
  chinese-english speech translation dataset}.

\end{thebibliography}

\label{lr:ref}
\bibliographystylelanguageresource{lrec-coling2024-natbib}
\bibliographylanguageresource{languageresource}

\end{document}